\documentclass[11pt]{article}
\usepackage{colacl}
\title{{\bf Introduction to the CoNLL-2002 Shared Task:
            Language-Independent Named Entity Recognition}}
\author{
\begin{tabular}{cc}
{\bf Erik F. Tjong Kim Sang}\\
CNTS - Language Technology Group\\
University of Antwerp\\
{\it erikt@uia.ua.ac.be}
\end{tabular}
}

\date{\today}

\begin{document}
\maketitle

\begin{abstract}
We describe the CoNLL-2002 shared task: language-independent named
entity recognition.
We give background information on the data sets and the evaluation
method, present a general overview of the systems that have taken part
in the task and discuss their performance.
\end{abstract}

\section{Introduction}

Named entities are phrases that contain the names of persons,
organizations, locations, times and quantities. Example:

\begin{quote}
  [PER Wolff ] , currently a journalist in [LOC Argentina ] , played
  with [PER Del Bosque ] in the final years of the seventies in [ORG
  Real Madrid ] .
\end{quote}

This sentence contains four named entities: 
{\it Wolff} and {\it Del Bosque} are persons,
{\it Argentina} is a location and 
{\it Real Madrid} is a organization.
The shared task of CoNLL-2002 concerns language-independent named
entity recognition. 
We will concentrate on four types of named entities: persons,
locations, organizations and names of miscellaneous entities that do
not belong to the previous three groups. 
The participants of the shared task have been offered training and
test data for two European languages: Spanish and Dutch.
They have used the data for developing a named-entity recognition
system that includes a machine learning component.
The organizers of the shared task were especially interested in 
approaches that make use of additional nonannotated data for improving
their performance.

\section{Data and Evaluation}

The CoNLL-2002 named entity data consists of six files covering two
languages: Spanish and Dutch\footnote{The data files are available
from http://lcg-www. uia.ac.be/conll2002/ner/}.
Each of the languages has a training file, a development file and a
test file.
The learning methods will be trained with the training data.
The development data can be used for tuning the parameters of the
learning methods.
When the best parameters are found, the method can be trained on the
training data and tested on the test data.
The results of the different learning methods on the test sets will 
be compared in the evaluation of the shared task.
The split between development data and test data has been chosen to
avoid that systems are being tuned to the test data.

All data files contain one word per line
with empty lines representing sentence boundaries.
Additionally each line contains a tag which states whether the word is
inside a named entity or not.
The tag also encodes the type of named entity.
Here is a part of the example sentence:

\begin{quote}
\begin{tabular}{rl}
Wolff      & B-PER \\
,          & O \\
currently  & O \\
a          & O \\
journalist & O \\
in         & O \\
Argentina  & B-LOC \\
,          & O \\
played     & O \\
with       & O \\
Del        & B-PER \\
Bosque     & I-PER \\
\end{tabular}
\end{quote}

Words tagged with O are outside of named entities.
The B-XXX tag is used for the first word in a named entity of type XXX
and I-XXX is used for all other words in named entities of type XXX.
The data contains entities of four types:
persons (PER),
organizations (ORG),
locations (LOC) and
miscellaneous names (MISC).
The tagging scheme is a variant of the IOB scheme originally put
forward by Ramshaw and Marcus \shortcite{ramshaw95}.
We assume that named entities are non-recursive and non-overlapping.
In case a named entity is embedded in another named entity usually
only the top level entity will be marked.

The Spanish data is a collection of news wire articles made available
by the Spanish EFE News Agency.
The articles are from May 2000.  
The annotation was carried out by the TALP Research
Center\footnote{http://www.talp.upc.es/} of the Technical University
of Catalonia (UPC) and the Center of Language and Computation
(CLiC\footnote{http://clic.fil.ub.es/}) of the University of Barcelona
(UB), and funded by the European Commission through the NAMIC project
(IST-1999-12392).
The data contains words and entity tags only.
The training, development and test data files contain 273037, 54837
and 53049 lines respectively.

The Dutch data consist of four editions of the Belgian newspaper
"De Morgen" of 2000 (June 2, July 1, August 1 and September 1).
The data was annotated as a part of the Atranos
project\footnote{http://atranos.esat.kuleuven.ac.be/} at the
University of Antwerp in Belgium, Europe.
The annotator has followed the MITRE and SAIC guidelines for named
entity recognition \cite{chinchor99} as well as possible.
The data consists of words, entity tags and part-of-speech tags which
have been derived by a Dutch part-of-speech tagger
\cite{daelemans96}.
Additionally the article boundaries in the text have been marked
explicitly with lines containing the tag -DOCSTART-.
The training, development and test data files contain 218737, 40656
and 74189 lines respectively.

The performance in this task is measured with F$_{\beta=1}$ rate which
is equal to ($\beta^2$+1)*precision*recall /
($\beta^2$*precision+recall) with $\beta$=1
\cite{vanrijsbergen75}.
Precision is the percentage of named entities found by the learning
system that are correct.
Recall is the percentage of named entities present in the corpus that
are found by the system.
A named entity is correct only if it is an exact match of the
corresponding entity in the data file.

\section{Results}

Twelve systems have participated in this shared task.
The results for the test sets for Spanish and Dutch can be found in
Table \ref{tab-results}.
A baseline rate was computed for both sets.
It was produced by a system which only identified entities which had a
unique class in the training data.
If a phrase was part of more than one entity, the system would select
the longest one.
All systems that participated in the shared task have outperformed
the baseline system.

McNamee and Mayfield \shortcite{mcnamee2002} have applied support
vector machines to the data of the shared task.
Their system used many binary features for representing words
(almost 9000).
They have evaluated different parameter settings of the system and
have selected a cascaded approach in which first entity boundaries
were predicted and then entity classes
(Spanish test set: F$_{\beta=1}$=60.97;
Dutch test set: F$_{\beta=1}$=59.52).

Black and Vasilakopoulos \shortcite{black2002} have evaluated two
approaches to the shared task.
The first was a transformation-based method which generated in rules
in a single pass rather than in many passes.
The second method was a decision tree method.
They found that the transformation-based method consistently
outperformed the decision trees
(Spanish test set: F$_{\beta=1}$=67.49;
Dutch test set: F$_{\beta=1}$=56.43)

Tsukamoto, Mitsuishi and Sassano \shortcite{tsukamoto2002}
used a stacked AdaBoost classifier for finding named entities.
They found that cascading classifiers helped improved performance.
Their final system consisted of a cascade of five learners each of
which performed 10,000 boosting rounds
(Spanish test set: F$_{\beta=1}$=71.49;
Dutch test set: F$_{\beta=1}$=60.93)

Malouf \shortcite{malouf2002} tested different models with the shared
task data: a statistical baseline model, a Hidden Markov Model and
maximum entropy models with different features.
The latter proved to perform best.
The maximum entropy models benefited from extra feature which 
encoded capitalization information, positional information and
information about the current word being part of a person name
earlier in the text.
However, incorporating a list of person names in the training process
did not help
(Spanish test set: F$_{\beta=1}$=73.66;
Dutch test set: F$_{\beta=1}$=68.08)

Jansche \shortcite{jansche2002}
employed a first-order Markov model as a named entity recognizer.
His system used two separate passes, one for extracting entity
boundaries and one for classifying entities.
He evaluated different features in both subprocesses.
The categorization process was trained separately from the extraction
process but that did not seem to have harmed overall performance
(Spanish test set: F$_{\beta=1}$=73.89;
Dutch test set: F$_{\beta=1}$=69.68)

Patrick, Whitelaw and Munro \shortcite{patrick2002}
present SLINERC, a language-independent named entity recognizer.
The system uses tries as well as character n-grams for encoding
word-internal and contextual information.
Additionally, it relies on lists of entities which have been 
compiled from the training data.
The overall system consists of six stages, three regarding entity
recognition and three for entity categorization.
Stages use the output of previous stages for obtaining an 
improved performance
(Spanish test set: F$_{\beta=1}$=73.92;
Dutch test set: F$_{\beta=1}$=71.36)

Tjong Kim Sang \shortcite{tjongkimsang2002} 
has applied a memory-based learner to the data of the shared task.
He used a two-stage processing strategy as well: first identifying 
entities and then classifying them.
Apart from the base classifier, his system made use of three extra
techniques for boosting performance: cascading classifiers (stacking),
feature selection and system combination.
Each of these techniques were shown to be useful
(Spanish test set: F$_{\beta=1}$=75.78;
Dutch test set: F$_{\beta=1}$=70.67).

Burger, Henderson and Morgan \shortcite{burger2002} have evaluated
three approaches to finding named entities.
They started with a baseline system which consisted of an HMM-based
phrase tagger.
They gave the tagger access to a list of approximately 250,000 named
entities and the performance improved.
After this several smoothed word classes derived from the available data
were incorporated into the training process.
The system performed better with the derived word lists than with the
external named entity lists
(Spanish test set: F$_{\beta=1}$=75.78;
Dutch test set: F$_{\beta=1}$=72.57).

Cucerzan and Yarowsky \shortcite{cucerzan2002}
approached the shared task by using word-internal and contextual
information stored in character-based tries.
Their system obtained good results by using part-of-speech tag
information and employing the {\it one sense per discourse} principle.
The authors expect a performance increase when the system has access
to external entity lists but have not presented the results of this in
detail
(Spanish test set: F$_{\beta=1}$=77.15;
Dutch test set: F$_{\beta=1}$=72.31).

Wu, Ngai, Carpuat, Larsen and Yang \shortcite{wu2002}
have applied AdaBoost.MH to the shared task data and compared the
performance with that of a maximum entropy-based named entity tagger.
Their system used lexical and part-of-speech information, contextual
and word-internal clues, capitalization information, knowledge
about entity classes of previous occurrences of words and a small
external list of named entity words.
The boosting techniques operated on decision stumps, 
decision trees of depth one.
They outperformed the maximum entropy-based named entity tagger
(Spanish test set: F$_{\beta=1}$=76.61;
Dutch test set: F$_{\beta=1}$=75.36).

Florian \shortcite{florian2002} employed three stacked learners for
named entity recognition:
transformation-based learning for obtaining base-level non-typed named
entities, Snow for improving the quality of these entities and the
forward-backward algorithm for finding categories for the named
entities.
The combination of the three algorithms showed a substantially
improved performance when compared with a single algorithm and an
algorithm pair
(Spanish test set: F$_{\beta=1}$=79.05;
Dutch test set: F$_{\beta=1}$=74.99).

Carreras, M{\`a}rquez and Padr\'o \shortcite{carreras2002}
have approached the shared task by using AdaBoost applied to
fixed-depth decision trees. 
Their system used many different input features contextual
information, word-internal clues, previous entity classes,
part-of-speech tags (Dutch only) and external word lists 
(Spanish only).
It processed the data in two stages: first entity recognition and 
then classification.
Their system obtained the best results in this shared task for both
the Spanish and Dutch test data sets
(Spanish test set: F$_{\beta=1}$=81.39;
Dutch test set: F$_{\beta=1}$=77.05).

\section{Concluding Remarks}

We have described the CoNLL-2002 shared task: language-independent
named entity recognition.
Twelve different systems have been applied to data covering two Western
European languages: Spanish and Dutch.
A boosted decision tree method obtained the best performance on both
data sets \cite{carreras2002}.

\section*{Acknowledgements}

Tjong Kim Sang is supported by IWT STWW as a researcher in the ATRANOS 
project.

\small

\bibliographystyle{acl}

\normalsize

\begin{table}[t]
\begin{center}
\begin{tabular}{|l|c|c|c|}\cline{2-4}
\multicolumn{1}{l|}{Spanish test}
                     & precision & recall  & F$_{\beta=1}$ \\\hline
Carreras et al.      & 81.38\%   & 81.40\% & 81.39 \\
Florian              & 78.70\%   & 79.40\% & 79.05 \\
Cucerzan et al.      & 78.19\%   & 76.14\% & 77.15 \\
Wu et al.            & 75.85\%   & 77.38\% & 76.61 \\
Burger et al.        & 74.19\%   & 77.44\% & 75.78 \\
Tjong Kim Sang       & 76.00\%   & 75.55\% & 75.78 \\
Patrick et al.       & 74.32\%   & 73.52\% & 73.92 \\
Jansche              & 74.03\%   & 73.76\% & 73.89 \\
Malouf               & 73.93\%   & 73.39\% & 73.66 \\
Tsukamoto            & 69.04\%   & 74.12\% & 71.49 \\
Black et al.         & 68.78\%   & 66.24\% & 67.49 \\
McNamee et al.       & 56.28\%   & 66.51\% & 60.97 \\\hline
baseline$^5$         & 26.27\%   & 56.48\% & 35.86 \\\hline
\multicolumn{4}{c}{}\\\cline{2-4}
\multicolumn{1}{l|}{Dutch test}
                     & precision & recall  & F$_{\beta=1}$ \\\hline
Carreras et al.      & 77.83\%   & 76.29\% & 77.05 \\
Wu et al.            & 76.95\%   & 73.83\% & 75.36 \\
Florian              & 75.10\%   & 74.89\% & 74.99 \\
Burger et al.        & 72.69\%   & 72.45\% & 72.57 \\
Cucerzan et al.      & 73.03\%   & 71.62\% & 72.31 \\
Patrick et al.       & 74.01\%   & 68.90\% & 71.36 \\
Tjong Kim Sang       & 72.56\%   & 68.88\% & 70.67 \\
Jansche              & 70.11\%   & 69.26\% & 69.68 \\
Malouf               & 70.88\%   & 65.50\% & 68.08 \\
Tsukamoto            & 57.33\%   & 65.02\% & 60.93 \\
McNamee et al.       & 56.22\%   & 63.24\% & 59.52 \\
Black et al.         & 62.12\%   & 51.69\% & 56.43 \\\hline
baseline$^5$         & 64.38\%   & 45.19\% & 53.10 \\\hline
\end{tabular}
\end{center}
\caption{
Overall precision, recall and F$_{\beta=1}$ rates obtained by the
twelve participating systems on the test data sets for the two
languages in the CoNLL-2002 shared task.
}
\label{tab-results}
\end{table}

\footnotetext[5]{
Due to some harmful annotation errors in the training data,
the baseline system performs less well than expected.
Without the errors, the baseline F$_{\beta=1}$ rates would have been 
62.49 for Spanish and 57.59 for Dutch.}

\end{document}